%% file: main.tex
\documentclass{article} 
\usepackage{iclr2027_conference,times}

\input{math_commands.tex}

\usepackage{hyperref}
\usepackage{url}
\usepackage{graphicx}
\usepackage{subcaption} 
\usepackage{booktabs} 

\title{DISRQAD: Diffusion Image Super-Resolution Quality Assessment Dataset and Benchmark}

\author{
\parbox[t]{0.46\textwidth}{\centering\normalfont
Nikita Kukuzei\thanks{Corresponding author:
\texttt{kukuzeyna@my.msu.ru}.}\\
Lomonosov Moscow State University\\
Moscow, Russia
}
\And
\parbox[t]{0.46\textwidth}{\centering\normalfont
Artem Borisov\\
Lomonosov Moscow State University\\
Moscow, Russia
}
\AND
\parbox[t]{0.46\textwidth}{\centering\normalfont
Evgeney Bogatyrev\\
MSU Institute for AI, Lomonosov Moscow State University\\
Moscow, Russia
}
\And
\parbox[t]{0.46\textwidth}{\centering\normalfont
Khaled Abud\\
MSU Institute for AI, Lomonosov Moscow State University\\
Moscow, Russia
}
\AND
\parbox[t]{0.46\textwidth}{\centering\normalfont
Egor Chistov\\
Lomonosov Moscow State University\\
Moscow, Russia
}
\And
\parbox[t]{0.46\textwidth}{\centering\normalfont
Dmitriy Vatolin\\
MSU Institute for AI, Lomonosov Moscow State University\\
Moscow, Russia
}
}

\iclrfinalcopy 

\begin{document}

\maketitle
\lhead{}

\begin{abstract}
Diffusion-based image super-resolution (SR) can create visually plausible detail that is not supported by the low-resolution input. We introduce DISRQAD, a subjective-quality dataset and diagnostic benchmark for this setting. It contains mean opinion scores (MOS) for 14,000 SR outputs from ten diffusion and four non-diffusion methods, spanning four low-resolution degradation conditions and $\times2$/$\times4$ upscaling. We evaluate 51 standard full-reference and no-reference metric configurations and 11 adapted variants. Agreement with MOS is substantially weaker on diffusion outputs: the strongest standard no-reference baseline reaches 0.431 SRCC on diffusion SR versus 0.813 on non-diffusion SR. As a case study in benchmark use, a pruned and distilled Q-ReAlign-mini student reaches 0.496 SRCC on diffusion SR. DISRQAD measures perceived output quality, not faithfulness to the input; it enables analysis of metric behavior across generator families and input conditions. Our findings reveal a substantial gap in the assessment of diffusion-based SR and provide a basis for developing quality models sensitive to diffusion-specific artifacts.
\end{abstract}


\section{Introduction}

Super-resolution (SR) reconstructs high-resolution (HR) images from low-resolution (LR) inputs, with deep neural networks driving recent progress~\cite{invsr,codsr}. However, SR methods can alter perceptually salient structures and introduce artifacts such as oversharpening, ringing, and unnatural textures~\cite{xie2023desra,srprominence}. Reliable image quality assessment (IQA) is therefore needed to compare reconstructed images at scale and guide the development of SR methods~\cite{fgresq,afine}. This task becomes particularly challenging when generated details look plausible but do not match the underlying scene. A mismatch between metric scores and human judgments can therefore distort comparisons between SR methods, even when they are evaluated on the same inputs.

\begin{figure}[t]
    \centering
    \includegraphics[width=\linewidth]{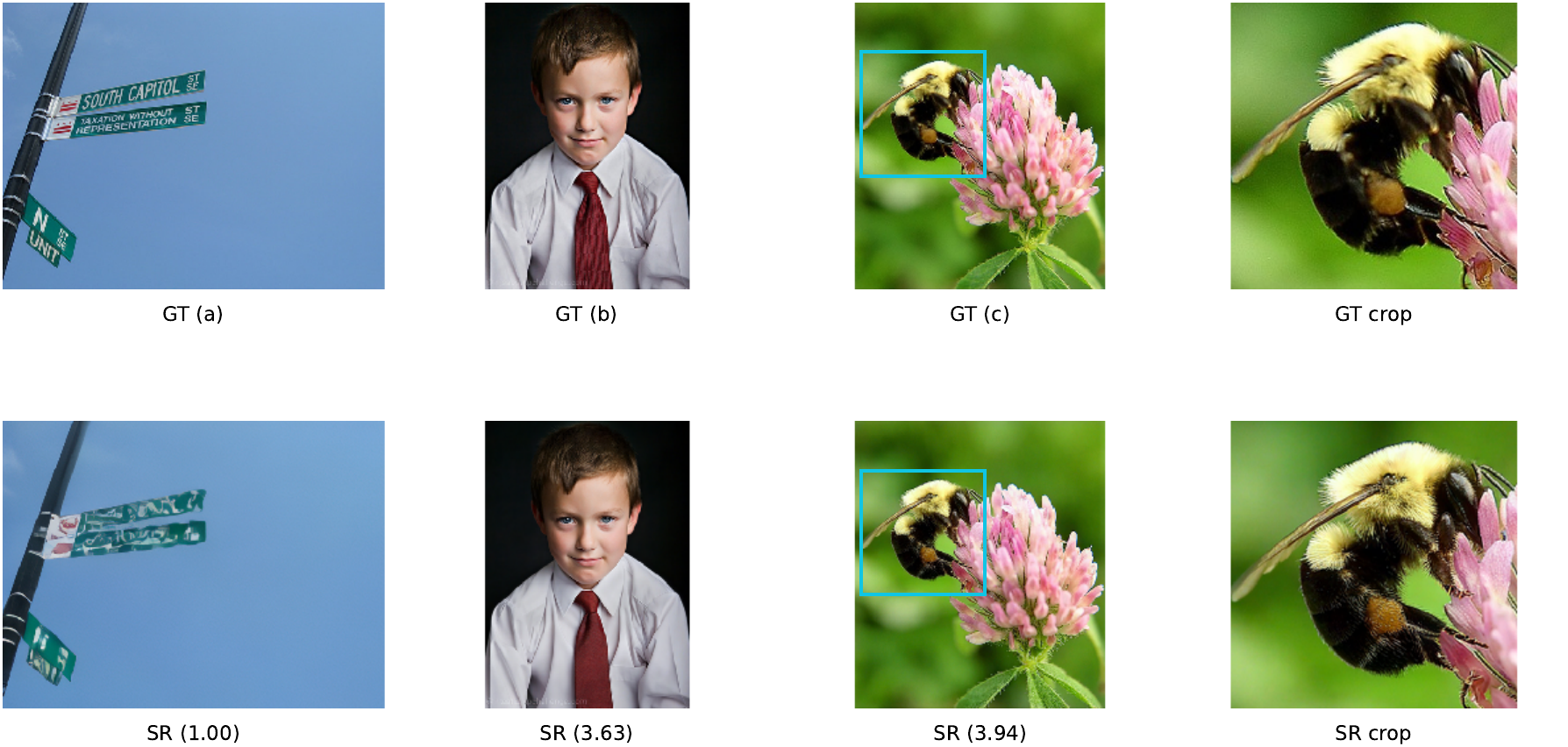}
    \caption{Comparison of diffusion-based super-resolution methods. The number in parentheses below each super-resolved image denotes its mean opinion score (MOS). (a) The super-resolved image exhibits significant artifacts. (b) The super-resolved image is perceptually comparable to the original image. (c) The super-resolved image is perceived as having higher visual quality than the original image.}
    \label{fig:mos_examples}
\end{figure}

In this paper, we use “diffusion-based SR” as an umbrella term for the evaluated generative restoration methods that employ diffusion- or flow-based priors. Diffusion-based Super-Resolution methods~\cite{resshift,seesr,supir} have demonstrated promising performance by generating images of higher quality. However, they still struggle to reconstruct plausible structures and textures, often hallucinating objects or modifying highly salient regions, such as human faces. Several works study perceptual artifacts generated by various SR models, including diffusion ones~\cite{srprominence, xie2023desra, srground}. Because a visually attractive output may differ from both the LR evidence and the HR reference, neither fidelity to the reference nor appearance alone provides an unambiguous quality target. Figure~\ref{fig:mos_examples} illustrates the range of perceived quality among diffusion-based SR outputs. In this work, we measure the visual quality of each output without showing raters its LR input or HR reference; fidelity to the underlying scene is a separate assessment target. This ambiguity may affect full-reference metrics, which penalize departures from the HR image, and no-reference metrics, which may reward plausible but unsupported detail. Human ratings are therefore needed to test how well existing IQA metrics rank generative SR outputs.

IQA metrics can be broadly divided into three categories: Full-Reference (FR) \cite{lpips,ssim,erqa}, Reduced-Reference (RR) \cite{Zheng_2021_ICCV}, and No-Reference (NR) \cite{qalign,arniqa,paq2piq}. FR metrics require access to an undistorted reference image, NR metrics estimate image quality using only the distorted image, whereas RR metrics rely on limited information extracted from the reference image. Recent studies~\cite{vcdb_vg} have shown that existing computational metrics exhibit only moderate agreement with human judgments. Furthermore, their limitations become even more pronounced when assessing super-resolved images~\cite{sisr}. Existing subjective studies that include diffusion SR cover relatively few generative methods~\cite{afine,isrgenqa}. It remains unclear whether the agreement between metric scores and human judgments is consistent across a broader range of generators and input conditions.

Therefore, one of the main objectives of this work is to determine which IQA metrics are most suitable for evaluating images produced by diffusion-based SR methods. We also investigate how well SR-specific IQA metrics perform on such data and analyze how fine-tuning on diffusion-generated images affects metric performance.

To evaluate IQA methods on generative SR, we first construct a diverse pool of LR inputs and outputs from ten diffusion and four non-diffusion SR methods. We then obtain MOS annotations for 10,000 diffusion and 4,000 non-diffusion outputs using more than 100,000 quality judgments. Finally, we compare 51 FR and NR IQA configurations against these human judgments and analyze the effects of SR-specific adaptation.

\textbf{Our main contributions are as follows:}
\begin{enumerate}
  \item \textbf{DISRQAD: a large-scale comprehensive dataset} of 10,000 diffusion-based SR images and 4,000 classic SR images with 205,578 judgments. HR images constitute a comprehensive dataset of 10036 semantically diverse and structurally complex high-quality images.
  \item \textbf{A comparative evaluation} of 51 IQA metrics of different types (FR and NR) and purposes, including SR-specific methods. We evaluated metrics from different families: traditional, CNN-based, and VLM-based.
  \item \textbf{Analysis and enhancement of representative IQA metrics:} we fine-tune representative metrics on diffusion-SR specific data to evaluate how SR-specific training helps to align metrics with human visual perception.
\end{enumerate}

\section{Related work}
\subsection{Subjective datasets for super-resolution}

Early SR-IQA datasets established human judgments as a basis for comparing reconstruction methods. CVIU-2017, QADS, and SISRSet vary algorithms and upscaling factors~\citep{nrqm,qads,sisr}; RealSRQ adds captured low-resolution inputs~\citep{realsrq}. SISAR extends coverage by inferring quality scores from a smaller set of direct ratings~\citep{sisar}. These datasets chiefly represent non-diffusion SR, where blur, ringing, and loss of detail are common quality concerns.

Recent studies also evaluate generative SR. SRIQA-Bench includes five diffusion methods and examines cases in which an enhanced image is preferred to its reference~\citep{afine}. ISRGen-QA provides direct mean opinion scores for five diffusion methods~\citep{isrgenqa}. Our benchmark places more diffusion generators alongside non-diffusion SR within one rating procedure. Table~\ref{tab:sr_dataset_comparison} summarizes the datasets and their annotation methods.

\input{tab_dataset_comparison.tex}

\subsection{IQA benchmarks}

Subjective protocols ask different questions. SBS180K records which of two SR outputs from the same input is preferred~\citep{neuralsbs}. FGRestore and LL-Bench combine such preferences with image-level scores across restoration tasks~\citep{fgresq,llbench}. They broaden task coverage; our analysis holds the task to SR and varies the generator family. In DISRQAD, raters score each SR output directly on a five-category scale. We average their ratings into MOS.

\citet{su2026rqi} document cases where FR and NR metrics disagree with pairwise human preferences on SR outputs that include diffusion. Agreement with image-level scores is commonly measured by Pearson (PLCC), Spearman (SRCC), or Kendall (KRCC) correlation~\citep{pearson1895,spearman1904,kendall1938}. We report signed SRCC and PLCC against direct MOS for non-diffusion and diffusion SR separately. This asks whether a metric that follows human ratings on non-diffusion SR also follows them on diffusion SR. Video-SR benchmarks such as VSRQAD~\citep{vsrqad} and \cite{herb2026accurate} also examine generative restoration, but temporal consistency and video-specific artifacts distinguish their annotation setting from our still-image benchmark.

Related work on video SR has compared quality metrics with human preferences for codec-compressed outputs~\cite{bogatyrev2024srcodec} and benchmarked real-time SR on streaming content~\cite{bogatyrev2026streamsr}.

\subsection{Image quality metrics}

Full-reference (FR) metrics compare an SR output with a high-resolution reference; no-reference (NR) metrics judge the output alone. In generative SR, synthesized texture may look natural but differ from the reference. FR metrics measure correspondence with that reference; NR metrics assess appearance without it.

\textbf{Full-reference metrics.} FR methods do not share a single notion of correspondence. PSNR and SSIM reward pixel or structural similarity~\citep{ssim}, whereas LPIPS and DISTS compare feature representations~\citep{lpips,dists}. ERQA focuses on restored edges~\citep{erqa}; A-FINE considers the case of an imperfect reference~\citep{afine}. Their assumptions matter when SR synthesizes plausible detail that does not reproduce the reference texture.

\textbf{No-reference metrics.} NR methods differ in their supervision. NIQE models natural-scene statistics without human scores~\citep{niqe}; MUSIQ learns from subjective ratings~\citep{musiq}, while Q-Align and DEQA use vision--language representations~\citep{qalign,deqa}. NeuralSBS learns SR preferences~\citep{neuralsbs}; FGResQ combines scores and rankings for restoration~\citep{fgresq}. These methods can assess appearance without an HR image, but cannot compare generated details with that image. Table~\ref{tab:metric_leaderboard} includes the full range of FR and NR metrics.

\section{DISRQAD Benchmark}
\label{sec:disrqad}

\begin{figure}[t]
  \centering
  \includegraphics[width=\linewidth]{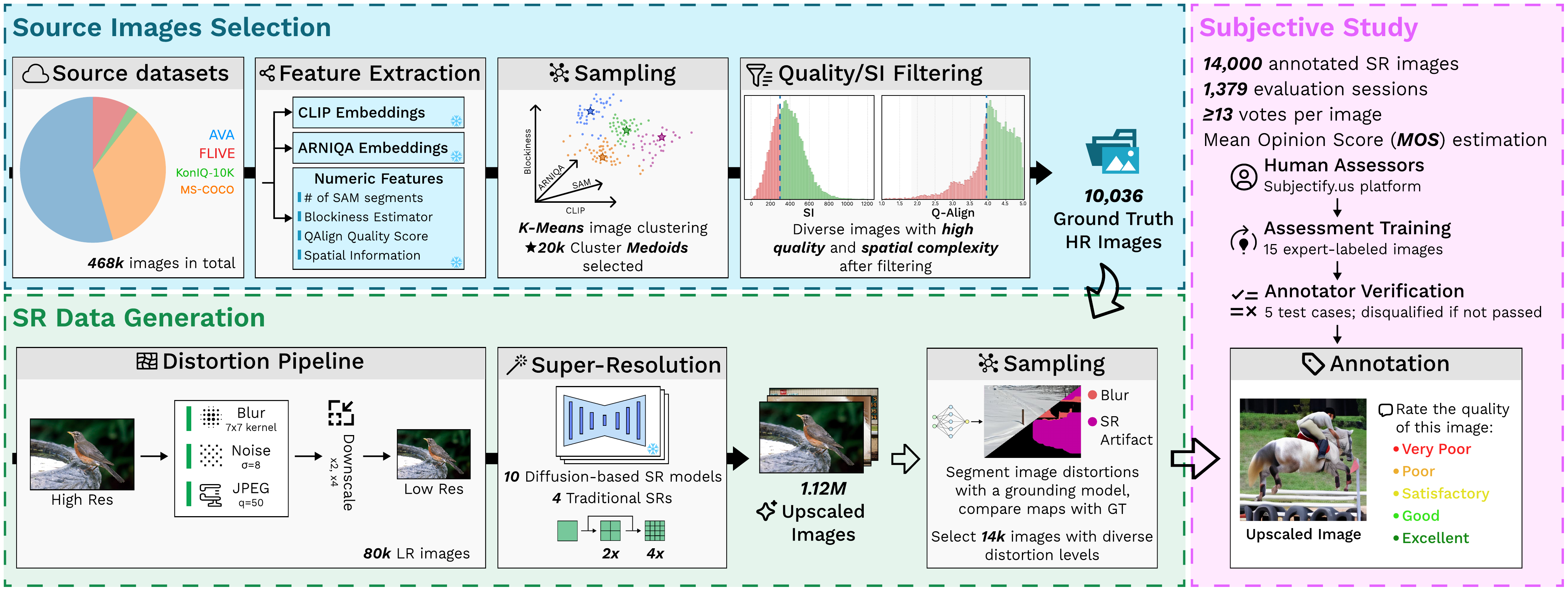}
  \caption{DISRQAD construction: source selection, LR condition generation, SR outputs, and subjective subset selection.}
  \label{fig:main_dataset_figure}
\end{figure}

Figure~\ref{fig:main_dataset_figure} summarizes the pipeline. We clustered more than 468,000 candidate images from KonIQ-10K, COCO, AVA, and FLIVE using structural and semantic features, then selected 10,036 sources using spatial-information (SI) and Q-Align thresholds. Q-Align filtered source-image quality before SR generation. The SI/Q-Align selection favors detailed, visually clean sources and excludes low-SI or low-Q-Align images from the rated benchmark. Appendix~\ref{app:construction} gives the features and thresholds.

We constructed four LR conditions per source and scale: resized inputs and inputs with blur, additive noise, or JPEG compression, at $\times2$ and $\times4$. These yield 80,288 LR images. We applied ten diffusion and four non-diffusion SR methods, then selected 10,000 and 4,000 outputs, respectively, by clustering output features. Each output received a direct MOS via subjective comparison. The open part of DISRQAD is available at \url{https://huggingface.co/datasets/visualprior/DISRQAD}. Further construction details appear in Appendix~\ref{app:construction}.

\subsection{Subjective target and protocol}
Our final target is perceived output quality, not reconstruction fidelity. Assessors viewed each SR output without its LR input or HR reference and rated it on a five-category scale from ``very poor'' to ``excellent.'' We averaged the ratings to obtain MOS. Thus, a visually attractive but scene-inconsistent image may receive a high score. 

A 50-output pilot compared pairwise preference, FR ratings, and NR ratings; its limited sample motivated the selected NR protocol but does not establish general equivalence between protocols. Training and control items screened annotators. The final annotation collected at least 13 ratings per output, yielding more than 205,000 judgments from more than 80 observers. In the pilot's vote-count analysis (Table~\ref{tab:different_votes_srocc}), the 13-rating MOS had SRCC 0.983 with the 20-rating baseline (95\% CI $[0.970,0.990]$); this measures agreement within that pilot, not uncertainty for every final MOS. Appendix~\ref{app:subjective} gives the protocol, categories, and pilot results.

\begin{table}[t]
    \centering
    \caption{SROCC between MOS computed from different numbers of ratings and the 20-rating baseline, with 95\% confidence intervals.}
    \label{tab:different_votes_srocc}
    \small
    \setlength{\tabcolsep}{7pt}
    \begin{tabular}{rrrr}
        \hline
        Ratings per image & SROCC & \multicolumn{2}{c}{95\% confidence interval} \\
                          &       & Lower bound & Upper bound \\
        \hline
        10 & 0.982547 & 0.969290 & 0.990110 \\
        11 & 0.978173 & 0.961658 & 0.987619 \\
        12 & 0.973870 & 0.954175 & 0.985165 \\
        \textbf{13} & \textbf{0.983084} & \textbf{0.970229} & \textbf{0.990415} \\
        14 & 0.983596 & 0.971124 & 0.990706 \\
        15 & 0.988575 & 0.979851 & 0.993534 \\
        16 & 0.991869 & 0.985641 & 0.995402 \\
        17 & 0.993483 & 0.988485 & 0.996316 \\
        18 & 0.997356 & 0.997356 & 0.998507 \\
        19 & 0.998654 & 0.997617 & 0.999240 \\
        \hline
    \end{tabular}
\end{table}

\section{Experiments}
\label{sec:experiments}

\subsection{Metric selection and evaluation}
We evaluate 51 standard FR and NR configurations, spanning conventional, learned, SR-oriented, and vision--language IQA, plus 11 adapted variants. Each standard configuration and its citation appear in Table~\ref{tab:metric_leaderboard}. We used each metric's default preprocessing; lower-is-better scores were sign-reversed. SRCC measures ranking agreement and PLCC measures linear agreement with MOS, without nonlinear fitting. Correlations are computed separately on 10,000 diffusion and 4,000 non-diffusion outputs. The MOS-annotated DISRQAD subsets are used for evaluation: the adapted baselines are not trained on their MOS. The 110-image SRIQA-Bench set used to select the student checkpoint is a validation set, not an independent test set.

\input{tab_metric_leaderboard}

\subsection{Benchmark results}

Agreement with MOS decreases substantially on diffusion SR. Across the 51 configurations in Table~\ref{tab:metric_leaderboard}, the median SRCC decreases from 0.65 on non-diffusion outputs to 0.27 on diffusion outputs, and the median PLCC decreases from 0.60 to 0.31. The decrease spans both FR and NR methods with widely different designs and supervision, rather than being confined to one metric family.

The strongest baselines narrow but do not close this gap. TOPIQ(FR) decreases from 0.81/0.79 SRCC/PLCC on non-diffusion SR to 0.40/0.42 on diffusion SR. Q-Align, a strong NR baseline, decreases from 0.81/0.76 to 0.42/0.49. The conventional FR metric FSIM also remains competitive on diffusion SR, reaching 0.39/0.40. These results show that neither access to an HR reference nor general-purpose learned quality representations alone ensures agreement with human judgments of diffusion SR. Q-Align's comparatively strong performance without SR-specific adaptation motivates the metric developed in Section~\ref{sec:efficient_metric}.

The fine-tuned examples in the leaderboard further show that SR adaptation is not uniformly beneficial. Fine-tuning LPIPS-AlexNet changes diffusion SRCC only from 0.28 to 0.29, while LPIPS-VGG decreases from 0.30 to 0.27. MUSIQ improves on diffusion SR from 0.27 to 0.31 but decreases on non-diffusion SR from 0.60 to 0.50. A complete comparison of the adapted variants is therefore reported separately: improvements in one SR family should not be treated as evidence of transfer to the other.

\section{Analysis by Condition and SR Adaptation}

\subsection{Performance by content and input condition}

\input{tab_metrics_on_content_types}

All results described below are presented in Table ~\ref{tab:metrics_on_content_types}. Metric performance differs substantially between classic and diffusion-based SR. On classic outputs, Q-Align reaches an overall SRCC of 0.813, while TOPIQ-FR reaches 0.811. Their correlations on diffusion outputs fall to 0.415 and 0.397, respectively. This gap extends beyond the leading methods: overall SRCC is higher on classic SR for 60 of the 62 evaluated metric variants. Thus, strong performance on classic SR provides limited evidence that a metric will rank diffusion outputs according to human judgments.

The degradation used to produce the low-resolution input also affects the results. For diffusion SR, the blur subset yields the highest SRCC for almost all of metrics. Upscaling factor and SR generator introduce further variation. On diffusion outputs, 52 of 62 metric variants have higher SRCC at ×4 than at ×2. The direction is less consistent on classic SR, where only 25 variants score higher at ×4. 


The difficulty of quality prediction also varies substantially among diffusion SR generators. PassionSR and ResShift yield the highest within-method SRCC for 36 and 20 of the 62 metric variants, respectively. For example, Q-Align reaches 0.714 on PassionSR, while distilled Q-ReAlign-mini reaches 0.666 on PassionSR and 0.613 on ResShift. DiT4SR presents the opposite pattern: it produces the lowest within-method SRCC for 38 metric variants. CODSR and PASD are also challenging, with median SRCCs across metric variants of 0.135 and 0.158. The gap across generators shows why results pooled over all diffusion outputs should be accompanied by within-method correlations.

The differences among classic SR methods are smaller and depend more on the metric. RealSR has the highest median within-method SRCC across metric variants, while the medians for SwinIR, HAT, and Real-ESRGAN lie lower. Q-Align performs particularly well on RealSR and HAT, whereas TOPIQ-FR leads on Real-ESRGAN and SwinIR. Real-ESRGAN gives the lowest SRCC for 25 of the 62 variants, yet it is not uniformly the hardest case. Thus, no single ranking of classic SR methods captures the behavior of every IQA metric.

Adaptation to diffusion SR improves some results but can reduce performance on classic outputs. Fine-tuning TOPIQ-FR raises its diffusion SRCC from 0.397 to 0.436, while its classic SRCC falls from 0.811 to 0.694. Distilled Q-ReAlign-mini achieves the highest overall diffusion SRCC (0.496) and retains a classic SRCC of 0.729. These findings motivate reporting results by SR family, generator, degradation, and scale alongside aggregate scores. 

\subsection{Performance gain after fine-tuning}

\input{tab_finetuned_metrics}

We fine-tuned five full-reference (FR) methods (AHIQ, LPIPS-AlexNet, LPIPS-VGG, PieAPP, and TOPIQ-FR) and five no-reference (NR) methods (ARNIQA, MUSIQ, LIQE, TOPIQ-NR, and Q-ReAlign-mini) on SRIQA-Bench.
Fine-tuning was performed under the following settings: a batch size of~8, 30~epochs, an initial learning rate of~\(1\!\times\!10^{-4}\), MSE loss against the MOS labels, and the Adam optimizer, with the best checkpoint selected according to SRCC.
LIQE required a lower learning rate (\(5\!\times\!10^{-6}\)) to ensure convergence.
Due to the large model size and rapid convergence of Q-ReAlign-mini, its maximum number of epochs was limited to~2.
All training was carried out using the IQA-PyTorch interface.

We measured the performance gain on the classic and diffusion tracks of our benchmark, as well as the decline in quality on non-SR distortions.
The effect of fine-tuning depends on the metric and the evaluation domain.
Table \ref{tab:metric_leaderboard_finetuned} shows the change in SRCC/PLCC metrics after fine-tuning.
For most metrics the effect is positive (the correlation improved on at least one track, and for 3~metrics the correlation improved on both tracks), but there are also negative examples.
Presumably, individual fine-tuning configurations need to be selected for them.

On non-SR distortions (the KonIQ-10K, KADID, and SPAQ datasets), the quality of the metrics dropped significantly.
Compared with the specialized pipeline of a hierarchically distilled metric, simple fine-tuning on the SR domain reduces the universality of the metric for image quality assessment across different domains.
Thus, improving quality on the diffusion SR domain requires the development of specialized approaches that do not neglect quality on classic SR and other types of distortions.

\section{Efficient SR-adapted metric}
\label{sec:efficient_metric}

Q-Align is a strong NR baseline but has 8.2B parameters. We use the 0.8B Q-ReAlign-mini~\citep{qrealign} as teacher and starting point, prune its language stack to obtain a 340M-parameter student, and combine a five-token quality head with a regression head. Hierarchical distillation on KonIQ-10K, KADID-10K, and SPAQ precedes SR adaptation on 990 SRIQA-Bench images with 248 general-IQA replay images. The remaining 110 SRIQA-Bench images are used for checkpoint selection. The architecture, losses, and training settings are in Appendix~\ref{app:student}.
The student retains Q-ReAlign-mini's visual encoder and reduces its 24 language modules to 12 by pruning two Gated DeltaNet modules from each four-module group; its vocabulary is restricted to 30 tokens. One head scores five ordered quality tokens and a second regresses a continuous score from the last hidden state. A learned convex combination gives the final prediction. This compact description makes the reported parameter and throughput changes interpretable; the layer matching, losses, and optimizer settings remain in Appendix~\ref{app:student}.
Figure~\ref{fig:qrealign_figure} summarizes the pruning and two-stage training.

\begin{figure}[t]
    \centering
    \includegraphics[width=\linewidth]{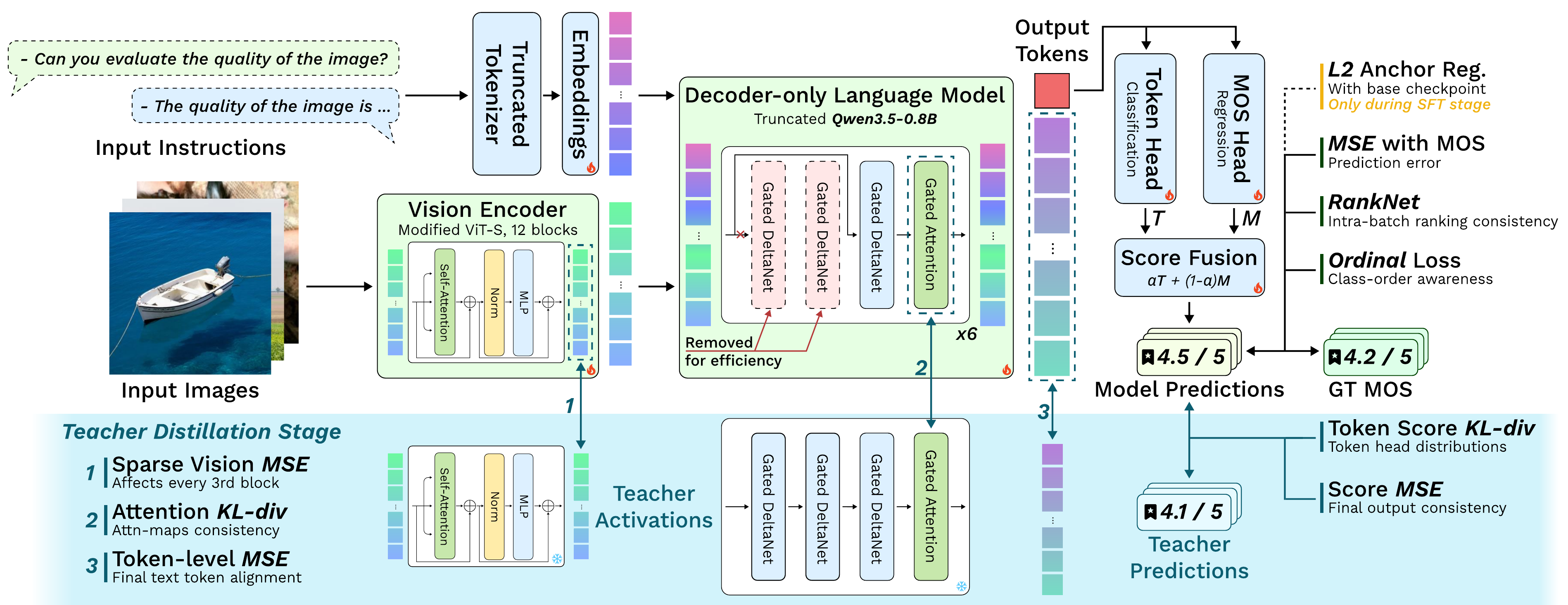}
    \caption{The process of Q-Align pruning and fine-tuning.}
    \label{fig:qrealign_figure}
\end{figure}

\input{tab_efficient_metric}

\section{Conclusion}

We introduced DISRQAD, a human-annotated benchmark of 14,000 SR outputs from ten diffusion and four non-diffusion methods, with 205,578 quality judgments. We evaluated 51 standard IQA configurations and 11 adapted variants against mean opinion scores for perceived output quality. For 60 of the 62 evaluated variants, SRCC was lower on diffusion outputs than on non-diffusion outputs. This pattern occurs across full-reference and no-reference metrics, while its magnitude varies with the SR generator, input degradation, and upscaling factor. Performance on conventional SR alone, therefore, provides an incomplete picture of a metric's suitability for generative SR.

As a case study, our pruned and distilled Q-ReAlign-mini student reaches an SRCC of 0.496 on diffusion SR and 0.729 on non-diffusion SR with 340M parameters. The broader fine-tuning results show that adaptation can improve agreement with human ratings, but its effects differ across metrics and SR families. Together, DISRQAD and these analyses provide a basis for developing and testing quality predictors for generative SR. The benchmark measures perceived output quality; assessing whether synthesized details remain faithful to the LR input is a complementary direction for future work.

\section*{AI disclosure}

We used generative AI tools for translation and language editing, assistance with drafting and revising code, and identification of relevant literature. The authors reviewed and edited AI-assisted material and take full responsibility for the code, analyses, results, and conclusions.

\section*{Reproducibility Statement}
Section~\ref{sec:disrqad} describes the construction of DISRQAD, the subjective quality target, and access to the released portion of the dataset. Section~\ref{sec:experiments} specifies the evaluated IQA configurations and correlation measures. Appendix~\ref{app:construction} details source selection, input degradations, SR methods, and output sampling; Appendix~\ref{app:subjective} documents the annotation protocol and pilot study; and Appendix~\ref{app:student} provides the student architecture and training settings.

\section*{Acknowledgments}
The research was carried out using the MSU-270 supercomputer of Lomonosov Moscow State University.

\bibliography{iclr2027_conference}
\bibliographystyle{iclr2027_conference}


\appendix

\section{Dataset construction}
\label{app:construction}
\subsection{Source selection}
Table~\ref{tab:feature_characteristics} lists the features used in source selection.

\begin{table}[ht]
\centering
\caption{Calculated image features and their descriptions.}
\label{tab:feature_characteristics}
\begin{tabular}{l l p{6cm}}
\hline
\textbf{Feature} & \textbf{Type} & \textbf{Description} \\
\hline
SI (Spatial Information) & Numeric & Characterizes the amount of detail and texture in the scene. \\
Blockiness~\cite{blockiness} & Numeric & Evaluates how pronounced the artificial boundaries between adjacent image blocks are. \\
SAM~\cite{sam} & Numeric & Number of objects detected by the model in the image. \\
Q-Align & Numeric & Expresses not only the visual quality of the image but also its aesthetic component, as it was trained on both tasks. \\
ARNIQA~\cite{arniqa} & Embedding & Reflects information about image quality and visual distortions. \\
CLIP~\cite{clip} & Embedding & Expresses semantic information about the image. \\
\hline
\end{tabular}
\end{table}

The source-selection portion of Figure~\ref{fig:main_dataset_figure} yields 10,036 selected images from the candidate pool. We collected images from KonIQ-10K, CoCo (train, test, val splits), AVA, FLIVE; resulting in more than 468,000 images. Then, SI (Spatial Information), blockiness \cite{blockiness}, SAM \cite{sam}, Q-Align numeric features, ARNIQA \cite{arniqa}, and CLIP \cite{clip} embeddings were calculated. SI characterizes the amount of detail and texture in the scene, blockiness evaluates how pronounced the artificial boundaries between adjacent image blocks are. As a feature of the SAM, we took the number of objects that the model detects in the image. Q-Align expresses not only the visual quality of the image, but also its aesthetic component, since it was trained on both of these tasks. If ARNIQA embeddings reflect information about image quality and visual distortions, then CLIP embeddings express semantic information about the image. 

\begin{figure}[t]
    \centering

    \begin{subfigure}[b]{0.48\linewidth}
        \centering
        \includegraphics[width=\linewidth]{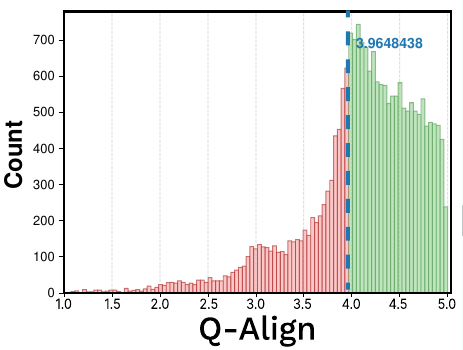}
        \caption{Distribution of Q-Align feature in medoid images.}
        \label{fig:qalign_dist}
    \end{subfigure}
    \hfill
    \begin{subfigure}[b]{0.48\linewidth}
        \centering
        \includegraphics[width=\linewidth]{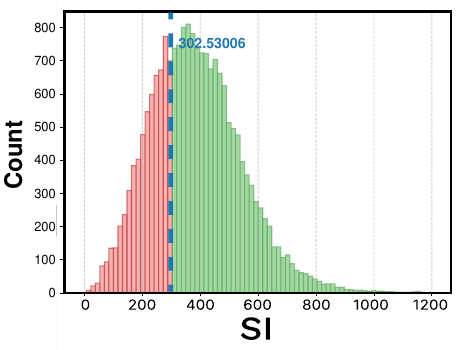}
        \caption{Distribution of SI feature in medoid images.}
        \label{fig:si_dist}
    \end{subfigure}

    \caption{Feature distributions in medoid images.}
    \label{fig:feature_distributions}
\end{figure}

We set a goal to take the most diverse images in terms of semantics and the number of objects presented, their dimensional features, natural and artificial distortions, while achieving maximum structural complexity and visual and aesthetic quality in order to evaluate SR methods in a highly competitive environment. We used K-means clustering to cluster images by blockiness, SAM, ARNIQA and CLIP. We identified 20,000 clusters from the entire image set. A medoid image was first determined for each cluster to serve as a representative exemplar. Subsequently, we applied a selection algorithm designed to identify images with the highest Q-Align and Spatial Information (SI) scores. Q-Align is a large multimodal model-based quality metric that has demonstrated state-of-the-art performance on image quality assessment benchmarks~\cite{qalign}, while SI quantifies the amount of spatial detail in an image based on the standard deviation of Sobel-filtered intensities. We employed a binning procedure to determine a percentage threshold \( T \) at which the intersection of the top \( T\% \) of images ranked independently by SI and Q-Align most closely approximated the target sample size. Using the resulting quantile, we fixed the Q-Align threshold and, from the remaining candidate pool, selected the maximum SI threshold such that the resulting number of images was greater than or equal to 10,000. This procedure yielded a final set of 10,036 images, characterized by thresholds of \( \text{SI} = 302.53006 \) and \( \text{Q-Align} = 3.9648438 \). The distributions of these features are presented in Figure~\ref{fig:feature_distributions}; its Q-Align and SI panels are shown in Figures~\ref{fig:qalign_dist} and~\ref{fig:si_dist}.

The 10,036 selected sources yield 80,288 LR images across four conditions and two scales. We apply next types of distortions: gaussian blur with quadratic kernel of 7 pixels and standard deviation 1.2, random noise with $X \sim \mathcal{N}(0, 8^2)$ and JPEG with quality 50\%. We also applied 2x and 4x downscaling using cubic interpolation. This gives eight LR images per source.  

\subsection{SR dataset construction}

We selected ten representative diffusion-based super-resolution methods (ResShift \cite{resshift}, SinSR \cite{sinsr}, SeeSR \cite{seesr}, SUPIR \cite{supir}, PASD, PASD-SDXL \cite{pasd}, InvSR \cite{invsr}, CODSR \cite{codsr}, DiT4SR \cite{dit4sr} and PassionSR \cite{passionsr}) to ensure broad coverage of the current design space of diffusion SR models. The benchmark includes methods based on different diffusion formulations, network architectures, conditioning strategies, and training paradigms. Specifically, the selected models range from early diffusion-based restoration approaches (e.g., ResShift) to recent latent diffusion and foundation-model-based methods (e.g., PASD-SDXL, InvSR, CODSR).

Thus, each method was applied to all images. If a model did not support configurable upscaling factors or lacked ×2 or ×4 scaling, we first upscaled the image to the smallest available scale exceeding the target resolution of reference image and then resized the output back to the target resolution using bicubic interpolation. Using this method, we obtained 802,880 SR images.

We also add non-diffusion SR models in our benchmark to compare performance of IQA metrics on images produced by both diffusion and classic SR approaches. Therefore, we included in the dataset images obtained using SwinIR, RealSR, HAT and Real-ESRGAN methods. Using these methods, we obtained 321,152 images.

Since resources for conducting subjective comparison are limited, we need to choose a small subset for comprehensive evaluation of IQA metrics. To include various types of distortions generated by SR methods in the dataset, we use an SR-grounding model \cite{srground} that generates segmentation maps for the different distortion types. We calculate class areas of every distortion and cw-ssim metric \cite{cw-ssim} between such maps as a reference and the corresponding SR images. We also calculate Q-Align \cite{qalign} and TOPIQ-FR \cite{topiq} because they have good correlation with human assessments, but the lowest correlation with each other according to \cite{vsrqad}. We applied K-means clustering to select 10,000 medoids from diffusion SR outputs and 4,000 from non-diffusion SR outputs separately. This second selection is based on output features; semantic-cluster coverage after selection has not been established.

\section{Subjective protocol and pilot}
\label{app:subjective}

We designed the subjective protocol to rate perceived output quality at a feasible annotation cost. NR MOS does not measure content fidelity to the LR input or HR reference. To select the protocol, we compared pairwise preferences with two forms of direct quality rating: full-reference (FR) and no-reference (NR).

\paragraph{Candidate protocols.}
In pairwise assessment, assessors select the better of two SR images, and the preferences are aggregated using the Bradley--Terry (BT) model~\cite{bradley1952rank}. Choosing between two images is generally easier for non-expert assessors than assigning an absolute quality score. We compared SR outputs derived from the same GT image. However, exhaustive pairwise comparison requires a number of judgments that grows quadratically with the number of outputs per reference. Moreover, BT scores estimated independently for different reference images are not directly comparable across references. We therefore used pairwise assessment as a baseline for selecting the rating protocol, rather than for annotating the entire dataset.

Direct quality ratings provide scores on a common scale, with an annotation cost that grows linearly with the number of images. Assessors assign each SR image a score from 1 to 5, with higher values indicating better quality, and the ratings are averaged to obtain its mean opinion score (MOS). Because assigning absolute scores can be difficult for non-experts, we evaluated both of the following protocols:
\begin{itemize}
    \item \textbf{Full-reference (FR).} Assessors view both the GT reference and the SR image. The five categories, corresponding to scores 1 through 5, are ``very poor quality compared to the reference,'' ``significantly worse than the reference,'' ``slightly worse than the reference,'' ``comparable to the reference,'' and ``better than the reference.''
    \item \textbf{No-reference (NR).} Assessors view only the SR image. The five categories, corresponding to scores 1 through 5, are ``very poor,'' ``poor,'' ``satisfactory,'' ``good,'' and ``excellent.''
\end{itemize}

\paragraph{Pilot study and quality control.}
We evaluated all three protocols on a pilot subset of 50 SR images spanning different distortions and quality levels: 5 outputs for each of 10 reference images. We collected at least 20 judgments per image for each MOS protocol and at least 20 judgments per pair for pairwise assessment. Pairwise comparisons were conducted on Subjectify.us, while MOS ratings were collected on Yandex Tasks.

\begin{figure*}[t]
    \centering
    \includegraphics[width=\textwidth]{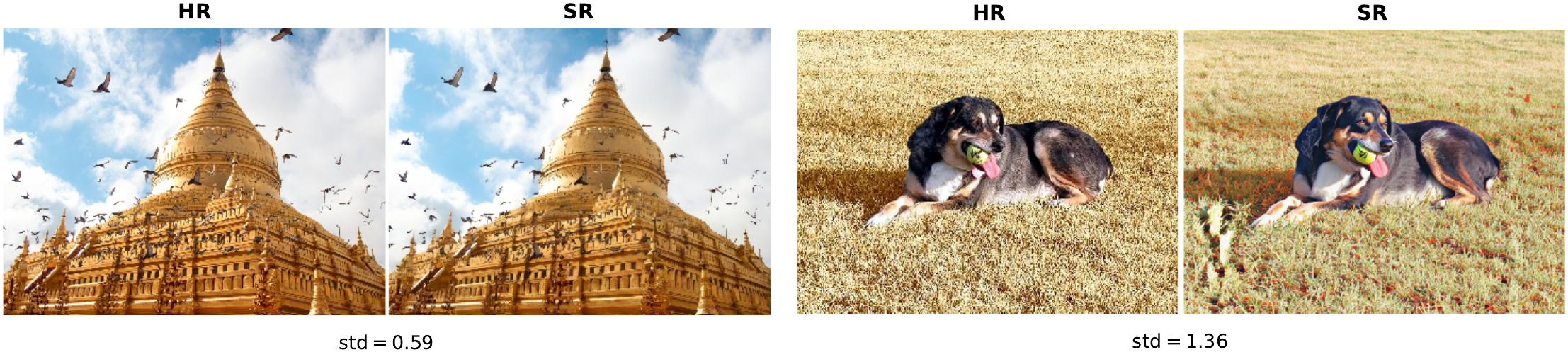}
    \caption{Examples of HR--SR image pairs filtered by the
    highest-standard-deviation criterion during in-lab annotation.
    This filtering was performed when selecting examples for the
    assessors' training and examination sets.}
    \label{fig:filtered_inlab}
\end{figure*}

We used control items with predetermined answers to screen out unreliable assessors. Only assessors who successfully completed the training were admitted to the MOS tasks. To construct the control items and training examples, we first conducted an in-lab study on 20 images spanning different distortions and quality levels. 9 experts independently rated each image using the NR protocol. We excluded the 5 images with the highest variance in expert ratings and used the remaining 15 images to prepare the training examples and control items. Examples of images filtered out during the in-lab annotation stage are shown in Figure~\ref{fig:filtered_inlab}. 

\begin{figure}[t]
    \centering
    \begin{minipage}[t]{0.48\linewidth}
        \centering
        \includegraphics[width=\linewidth]{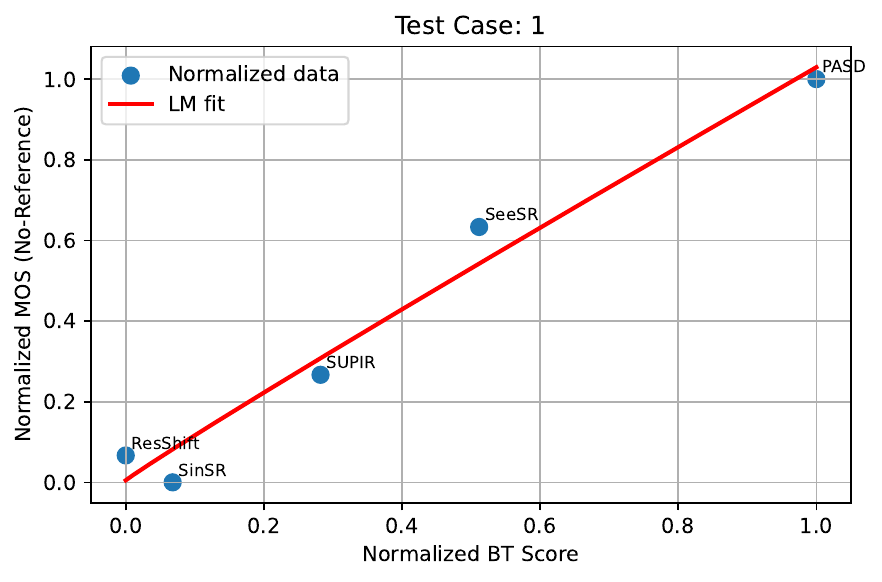}
        \par\smallskip
        {\small (1)}
    \end{minipage}\hfill
    \begin{minipage}[t]{0.48\linewidth}
        \centering
        \includegraphics[width=\linewidth]{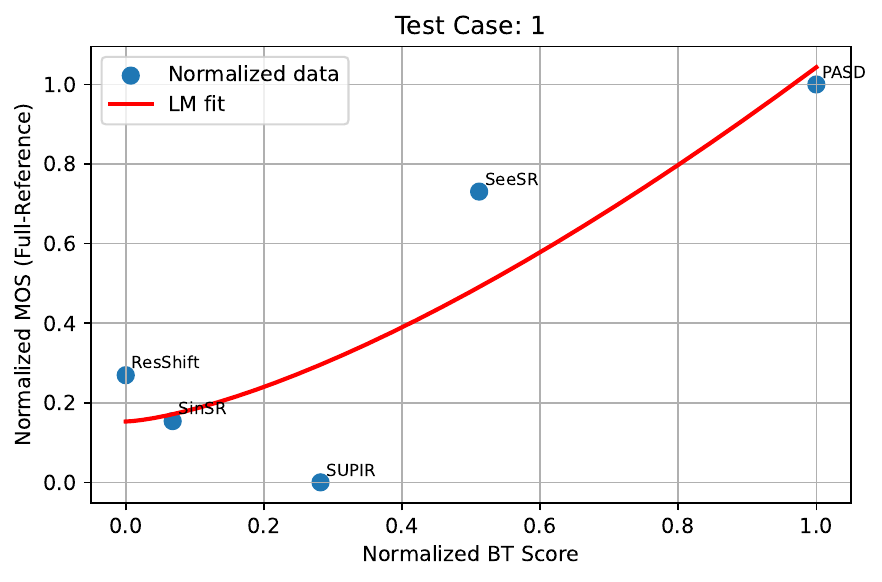}
        \par\smallskip
        {\small (2)}
    \end{minipage}
    \caption{LM-fitted relationships between MOS and pairwise Bradley--Terry scores for SR outputs associated with one reference image: (1) no-reference MOS and (2) full-reference MOS.}
    \label{fig:mos_lm_fitting_error}
\end{figure}

\begin{figure*}[t]
    \centering
    \includegraphics[width=0.98\textwidth]{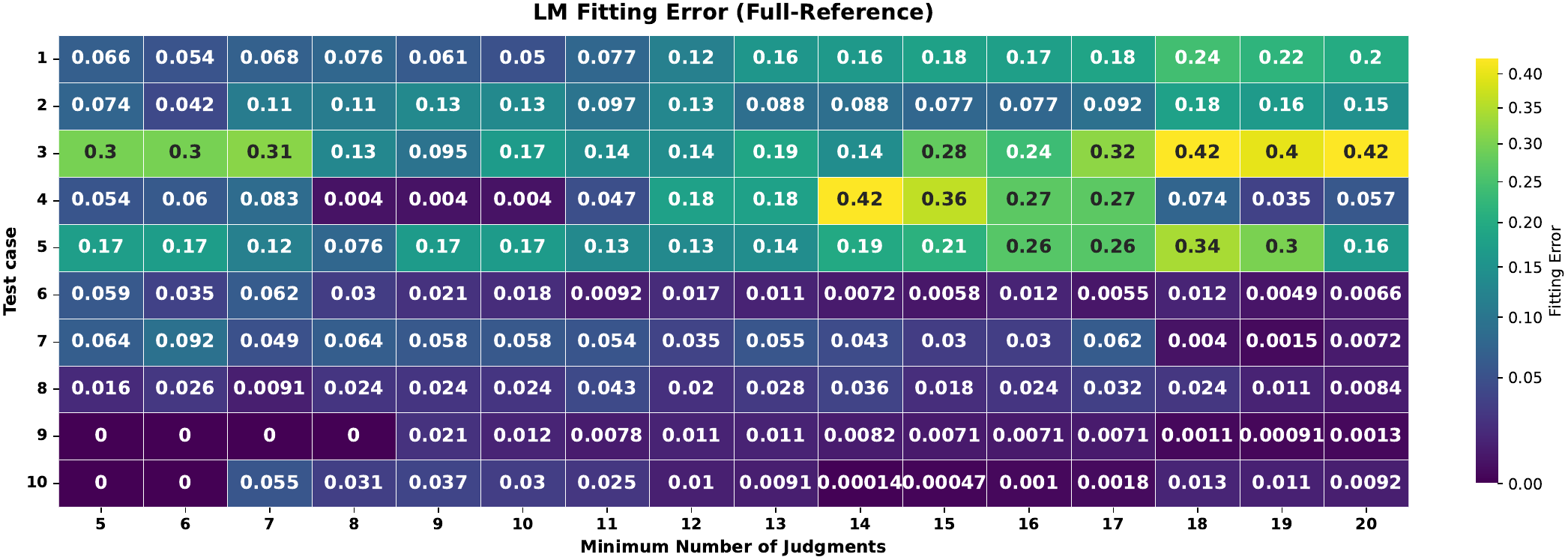}

    \vspace{0.5em}

    \includegraphics[width=0.98\textwidth]{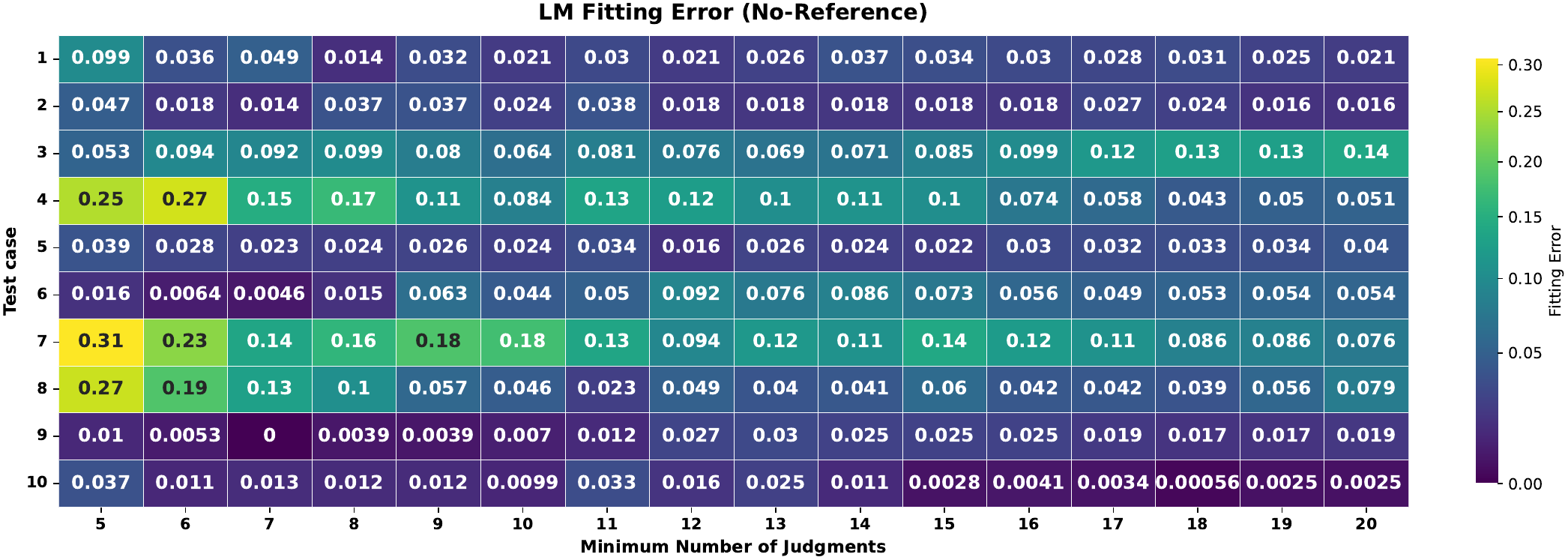}
    \caption{LM-fitting error between MOS and pairwise Bradley--Terry
    scores as a function of the minimum number of judgments per test
    case. The upper panel uses full-reference MOS, and the lower panel
    uses no-reference MOS. The panels use separate color scales.}
    \label{fig:lm_fitting_error_maps_long}
\end{figure*}

\paragraph{Selection of the MOS protocol.}
Following the strategy in~\cite{j87d-w690-24}, we evaluated how well each MOS protocol agreed with the pairwise scores after fitting a nonlinear mapping. We fitted the mapping separately for each reference image and each MOS protocol:
\begin{equation}
    (\hat{a}_i, \hat{b}_i, \hat{c}_i)
    = \mathop{\mathrm{arg\,min}}_{a_i,b_i,c_i}
    \sum_n \left(a_i x_{in}^{\,b_i} + c_i - z_{in}\right)^2,
    \label{eq:mos_bt_fitting}
\end{equation}
where $x_{in}$ is the MOS of the $n$th SR image associated with reference $i$, and $z_{in}$ is its BT score obtained from pairwise assessment. We estimated the parameters using the Levenberg--Marquardt (LM) algorithm~\cite{10.1007/BFb0067700} and assessed agreement using the average fitting error. In this 50-output pilot, NR MOS showed closer fitted agreement with pairwise scores than FR MOS, so we selected the NR protocol for subsequent experiments. This small pilot does not establish general superiority of NR ratings. The fitting objective in \eqref{eq:mos_bt_fitting} and Figure~\ref{fig:mos_lm_fitting_error} illustrate the fitted relationships for the SR outputs associated with one reference image. Figure ~\ref{fig:lm_fitting_error_maps_long} shows how the LM-fitting error for each set of SR images associated with a single HR reference varies with the number of judgments and the annotation protocol (no-reference or full-reference).

\paragraph{Number of ratings per image.}
To determine the annotation budget, we compared MOS computed from different numbers of ratings with baseline scores obtained from at least 20 ratings per image. Table~\ref{tab:different_votes_srocc} reports Spearman's rank-order correlation coefficient (SROCC) and the corresponding 95\% confidence intervals. With 13 ratings per image, the pilot table reports SROCC 0.983084 with a 95\% confidence interval of $[0.970229,\,0.990415]$. We selected a minimum of 13 ratings per image for the final annotation, recognizing that the pilot confidence interval does not establish the precision of each final MOS.

\paragraph{Final protocol.}
We annotated the entire dataset using NR MOS on the five-category scale described above, collecting at least 13 ratings per image. We retained the training examples and control items used in the pilot study.

\section{Student architecture and training}
\label{app:student}

Q-Align provides a strong NR baseline in Table~\ref{tab:metric_leaderboard}, but its 8.2B parameters make direct adaptation and deployment expensive. Directly fine-tuning the smaller 0.8B Q-ReAlign-mini checkpoint~\citep{qrealign} on SRIQA-Bench raises its diffusion SRCC from 0.341 to 0.450, indicating that a compact vision--language metric can benefit from SR-specific supervision. As shown in Figure~\ref{fig:qrealign_figure}, we construct a smaller student from Q-ReAlign-mini and train it in two stages: hierarchical distillation on general IQA data followed by adaptation to SR.

\subsection{Student architecture}
 
We initialize the student from the Q-ReAlign-mini 0.8B checkpoint. Like Q-Align, it receives an image and a fixed quality-assessment prompt. The student retains the visual encoder but reduces the language component of the Qwen3.5-VL backbone. Its 24 language modules form six repeated groups of three Gated DeltaNet modules followed by one full-attention module. We remove the first two Gated DeltaNet modules from each group, leaving 12 language modules in total, and restrict the text vocabulary to 30 tokens. These changes reduce the model to 340M parameters.

The model predicts quality through two heads. The token head converts the probabilities of five ordered quality tokens into a score $T$. The regression head applies a $1024\!\rightarrow\!256$ projection, GELU, dropout, a $256\!\rightarrow\!1$ projection, and a sigmoid to the last hidden state, producing a continuous estimate $M$. The final prediction is the convex combination $\hat{y}=\alpha T+(1-\alpha)M$, where $\alpha$ is learned jointly with the model. This preserves Q-Align's discrete quality representation while allowing direct regression to MOS.

\subsection{Hierarchical distillation and SR adaptation}

In the first stage, the unpruned Q-ReAlign-mini teacher provides supervision at several levels of the network. We match every third visual-layer output using mean squared error (MSE), match every corresponding language-layer output using MSE, and minimize the KL divergence between the teacher and student full-attention maps. At the prediction level, we match the teacher score using MSE and its distribution over the five quality tokens using KL divergence.

The student is simultaneously trained against subjective labels from KonIQ-10K~\citep{koniq10k}, KADID-10K~\citep{kadid10k}, and SPAQ~\citep{spaq}, using the dataset splits adopted by Q-Align. The label-based terms comprise MSE between $\hat{y}$ and MOS, an in-batch RankNet loss, and an ordinal loss over the levels $\{0,0.25,0.5,0.75,1\}$. When MOS falls between two adjacent levels, their target probabilities are obtained by linear interpolation; the ordinal loss is the mean squared distance between their cumulative distribution and that of the five predicted tokens. An additional MSE term encourages the two heads to produce consistent scores. In total, the first-stage objective contains nine terms, weighted as follows: MOS 1.0, ranking 0.3, ordinal 0.2, head consistency 0.05, teacher score 0.4, teacher tokens 0.5, text features 0.02, visual features 0.01, and attention 0.01. We train for eight epochs with an effective batch size of 8 (physical batch size 2) and select the checkpoint with the highest validation SRCC.

The second stage removes the teacher and adapts the student to SR. Its training set combines 990 SRIQA-Bench images with 248 images from the three general-IQA datasets. SRIQA-Bench assigns MOS to both SR outputs and the corresponding HR images. Its 1,100 rated images were split at the individual-image level. The objective retains the first four label-based terms and adds L2 regularization toward the first-stage checkpoint with weight $10^{-4}$. Both stages use AdamW with weight decay 0.01, 5\% linear warm-up followed by cosine decay, and learning rates $2\times10^{-5}$, $2\times10^{-6}$, and $10^{-4}$ for the language, vision, and head parameters, respectively. This stage runs for two epochs with an effective batch size of 8 (physical batch size 4). The remaining 110 SRIQA-Bench images are used as validation for checkpoint selection.

\subsection{Diffusion SR adaptation}

Starting from the SRIQA-Bench-adapted checkpoint, we further fine-tune the student on the open part of DISRQAD, following the same fine-tuning strategy used for SRIQA-Bench. We then evaluate it on the held-out private portion. On the 4,000 diffusion SR images in the private split, the model achieves an SRCC of 0.761 and a PLCC of 0.773. On the 1,600 non-diffusion SR images, it achieves an SRCC of 0.890 and a PLCC of 0.879. These results demonstrate that learning to predict perceived quality in the diffusion SR domain is feasible. The remaining performance gap relative to non-diffusion SR indicates that diffusion outputs pose distinct assessment challenges, motivating further domain-specific adaptation, specialized architectures, and systematic study.

\section{Limitations}
\label{app:limitations}

DISRQAD focuses on the perceived visual quality of SR outputs under a no-reference protocol. Faithfulness to the LR input and semantic preservation are complementary assessment targets. The benchmark emphasizes detailed, high-quality source images and controlled single-degradation inputs, so applying its findings to compound real-world degradations calls for separate evaluation.




\end{document}

%% file: math_commands.tex
\usepackage{amsmath,amsfonts,bm}

\def\eqref#1{equation~\ref{#1}}

\def\1{\bm{1}}

\DeclareMathAlphabet{\mathsfit}{\encodingdefault}{\sfdefault}{m}{sl}
\SetMathAlphabet{\mathsfit}{bold}{\encodingdefault}{\sfdefault}{bx}{n}



%% file: tab_dataset_comparison.tex
\newcommand{\tabmark}[1]{\rlap{$^{#1}$}}

\begin{table}[!h]
  \centering
  \caption{SR datasets with per-image scores. Ratings are absolute judgments; pairs are preferences. Sources are distinct LR inputs for SISAR and HR inputs otherwise.}
  \label{tab:sr_dataset_comparison}
  \footnotesize
  \setlength{\tabcolsep}{3pt}
  \begin{tabular}{@{}lrrrrl@{}}
    \toprule
    Dataset & Sources & \shortstack{SR\\outputs} & \shortstack{Non-diff.\\SR} & \shortstack{Diffusion\\SR} & \shortstack{Human\\input} \\
    \midrule
    CVIU-2017~\citep{nrqm} & 30 & 1,620 & 9 & 0 & Ratings \\
    QADS~\citep{qads} & 20 & 980 & 21 & 0 & Pairs \\
    SISRSet~\citep{sisr} & 15 & 360 & 8 & 0 & Pairs \\
    RealSRQ~\citep{realsrq} & 60 & 1,620 & 10 & 0 & Pairs \\
    SISAR~\citep{sisar} & 100 & 12,600\tabmark{\ast} & 10\tabmark{\ast} & 0 & Ratings \\
    SRIQA-Bench~\citep{afine} & 100 & 1,000 & 5 & 5 & Pairs \\
    ISRGen-QA~\citep{isrgenqa} & 19 & 720 & 10 & 5 & Ratings \\
    RQI study~\citep{su2026rqi} & 312 & 2,184 & 4 & 3 & Pairs \\
    \midrule
    \textbf{DISRQAD (ours)} & \textbf{10,036}\tabmark{\dagger} & \textbf{14,000} & 4 & \textbf{10} & Ratings \\
    \bottomrule
  \end{tabular}
  \begin{minipage}{\linewidth}
    \vspace{2pt}
    \scriptsize\raggedright
    $^\ast$ SISAR directly rates only 1,800 of its 12,600 outputs; the remaining 10,800 scores are inferred. Its method count comprises six SR algorithms and four combined operations.\\
    $^\dagger$ DISRQAD reports the full source pool before SR generation, not the number of unique sources represented in the 14,000 subjectively evaluated outputs.
  \end{minipage}
\end{table}

%% file: tab_metric_leaderboard.tex
\begin{table*}[t]
\centering
\caption{Overall signed correlations with MOS for 51 standard full-reference (FR) and no-reference (NR) IQA configurations on non-diffusion and diffusion SR. Rows are ordered by non-diffusion SRCC within each type. Best and second-best results within each type and column are shown in bold and underlined, respectively. Values are rounded to three decimal places.}
\label{tab:metric_leaderboard}
\begingroup
\footnotesize
\setlength{\tabcolsep}{4pt}
\renewcommand{\arraystretch}{0.88}
\begin{tabular}{@{}lccccc@{}}
\toprule
& & \multicolumn{2}{c}{Non-diffusion SR} & \multicolumn{2}{c}{Diffusion SR} \\
\cmidrule(lr){3-4}\cmidrule(lr){5-6}
Metric & Type & SRCC & PLCC & SRCC & PLCC \\
\midrule
TOPIQ(FR)~\citep{topiq} & FR & $\phantom{-}\mathbf{0.811}$ & $\phantom{-}\mathbf{0.789}$ & $\phantom{-}\mathbf{0.397}$ & $\phantom{-}\mathbf{0.417}$ \\
AFINE(FR)~\citep{afine} & FR & $\phantom{-}\underline{0.787}$ & $\phantom{-}\underline{0.738}$ & $\phantom{-}0.310$ & $\phantom{-}\underline{0.410}$ \\
VSI~\citep{vsi} & FR & $\phantom{-}0.765$ & $\phantom{-}0.663$ & $\phantom{-}0.355$ & $\phantom{-}0.366$ \\
GMSD~\citep{gmsd} & FR & $\phantom{-}0.765$ & $\phantom{-}0.719$ & $\phantom{-}0.317$ & $\phantom{-}0.325$ \\
LPIPS-VGG~\citep{lpips} & FR & $\phantom{-}0.763$ & $\phantom{-}0.719$ & $\phantom{-}0.300$ & $\phantom{-}0.329$ \\
FSIM~\citep{fsim} & FR & $\phantom{-}0.758$ & $\phantom{-}0.681$ & $\phantom{-}\underline{0.386}$ & $\phantom{-}0.405$ \\
DISTS~\citep{dists} & FR & $\phantom{-}0.751$ & $\phantom{-}0.695$ & $\phantom{-}0.232$ & $\phantom{-}0.283$ \\
DMM~\citep{dmm} & FR & $\phantom{-}0.739$ & $\phantom{-}0.466$ & $\phantom{-}0.309$ & $\phantom{-}0.308$ \\
MS-SSIM~\citep{ms-ssim} & FR & $\phantom{-}0.725$ & $\phantom{-}0.611$ & $\phantom{-}0.308$ & $\phantom{-}0.288$ \\
LPIPS-AlexNet~\citep{lpips} & FR & $\phantom{-}0.723$ & $\phantom{-}0.655$ & $\phantom{-}0.282$ & $\phantom{-}0.311$ \\
WaDIQaM(FR)~\citep{wadiqam} & FR & $\phantom{-}0.715$ & $\phantom{-}0.701$ & $\phantom{-}0.366$ & $\phantom{-}0.373$ \\
VIF~\citep{vif} & FR & $\phantom{-}0.687$ & $\phantom{-}0.669$ & $\phantom{-}0.344$ & $\phantom{-}0.349$ \\
MAD~\citep{mad} & FR & $\phantom{-}0.685$ & $\phantom{-}0.641$ & $\phantom{-}0.260$ & $\phantom{-}0.257$ \\
NLPD~\citep{nlpd} & FR & $\phantom{-}0.678$ & $\phantom{-}0.603$ & $\phantom{-}0.290$ & $\phantom{-}0.282$ \\
AHIQ~\citep{ahiq} & FR & $\phantom{-}0.640$ & $\phantom{-}0.594$ & $\phantom{-}0.291$ & $\phantom{-}0.322$ \\
PSNR & FR & $\phantom{-}0.630$ & $\phantom{-}0.621$ & $\phantom{-}0.257$ & $\phantom{-}0.259$ \\
SSIM~\citep{ssim} & FR & $\phantom{-}0.597$ & $\phantom{-}0.566$ & $\phantom{-}0.271$ & $\phantom{-}0.272$ \\
PieAPP~\citep{pieapp} & FR & $\phantom{-}0.567$ & $\phantom{-}0.451$ & $\phantom{-}0.214$ & $\phantom{-}0.269$ \\
CW-SSIM~\citep{cw-ssim} & FR & $\phantom{-}0.504$ & $\phantom{-}0.437$ & $\phantom{-}0.183$ & $\phantom{-}0.156$ \\
ERQA~\citep{erqa} & FR & $\phantom{-}0.493$ & $\phantom{-}0.388$ & $\phantom{-}0.244$ & $\phantom{-}0.289$ \\
CKDN & FR & $\phantom{-}0.093$ & $\phantom{-}0.173$ & $\phantom{-}0.334$ & $\phantom{-}0.373$ \\
\midrule
Q-Align~\citep{qalign} & NR & $\phantom{-}\mathbf{0.813}$ & $\phantom{-}\mathbf{0.758}$ & $\phantom{-}\underline{0.415}$ & $\phantom{-}\mathbf{0.487}$ \\
Q-ReAlign-lite & NR & $\phantom{-}\underline{0.772}$ & $\phantom{-}0.719$ & $\phantom{-}0.405$ & $\phantom{-}0.458$ \\
DEQA~\citep{deqa} & NR & $\phantom{-}0.770$ & $\phantom{-}\underline{0.728}$ & $\phantom{-}0.324$ & $\phantom{-}0.392$ \\
Q-ReAlign-pro & NR & $\phantom{-}0.760$ & $\phantom{-}0.716$ & $\phantom{-}\mathbf{0.431}$ & $\phantom{-}\underline{0.469}$ \\
LIQE~\citep{liqe} & NR & $\phantom{-}0.719$ & $\phantom{-}0.684$ & $\phantom{-}0.279$ & $\phantom{-}0.342$ \\
NIMA~\citep{nima} & NR & $\phantom{-}0.702$ & $\phantom{-}0.671$ & $\phantom{-}0.329$ & $\phantom{-}0.357$ \\
HyperIQA~\citep{hyperiqa} & NR & $\phantom{-}0.688$ & $\phantom{-}0.662$ & $\phantom{-}0.208$ & $\phantom{-}0.306$ \\
Q-ReAlign-mini & NR & $\phantom{-}0.683$ & $\phantom{-}0.597$ & $\phantom{-}0.341$ & $\phantom{-}0.393$ \\
TRIQA~\citep{triqa} & NR & $\phantom{-}0.678$ & $\phantom{-}0.636$ & $\phantom{-}0.385$ & $\phantom{-}0.438$ \\
CLIPIQA~\citep{clipiqa} & NR & $\phantom{-}0.670$ & $\phantom{-}0.643$ & $\phantom{-}0.332$ & $\phantom{-}0.370$ \\
TOPIQ(NR)~\citep{topiq} & NR & $\phantom{-}0.668$ & $\phantom{-}0.623$ & $\phantom{-}0.256$ & $\phantom{-}0.344$ \\
MACLIP~\citep{maclip} & NR & $\phantom{-}0.651$ & $\phantom{-}0.625$ & $\phantom{-}0.274$ & $\phantom{-}0.317$ \\
DBCNN~\citep{dbcnn} & NR & $\phantom{-}0.642$ & $\phantom{-}0.601$ & $\phantom{-}0.225$ & $\phantom{-}0.315$ \\
MANIQA~\citep{maniqa} & NR & $\phantom{-}0.629$ & $\phantom{-}0.602$ & $\phantom{-}0.313$ & $\phantom{-}0.348$ \\
AFINE(NR)~\citep{afine} & NR & $\phantom{-}0.623$ & $\phantom{-}0.561$ & $-0.001$ & $\phantom{-}0.008$ \\
MUSIQ~\citep{musiq} & NR & $\phantom{-}0.605$ & $\phantom{-}0.501$ & $\phantom{-}0.273$ & $\phantom{-}0.332$ \\
FGResQ~\citep{fgresq} & NR & $\phantom{-}0.593$ & $\phantom{-}0.560$ & $\phantom{-}0.006$ & $\phantom{-}0.183$ \\
QualiCLIP~\citep{qualiclip} & NR & $\phantom{-}0.590$ & $\phantom{-}0.562$ & $\phantom{-}0.032$ & $\phantom{-}0.121$ \\
TReS~\citep{tres} & NR & $\phantom{-}0.567$ & $\phantom{-}0.482$ & $\phantom{-}0.158$ & $\phantom{-}0.241$ \\
CNNIQA~\citep{cnniqa} & NR & $\phantom{-}0.566$ & $\phantom{-}0.416$ & $\phantom{-}0.159$ & $\phantom{-}0.232$ \\
WaDIQaM(NR)~\citep{wadiqam} & NR & $\phantom{-}0.558$ & $\phantom{-}0.506$ & $\phantom{-}0.122$ & $\phantom{-}0.179$ \\
ARNIQA~\citep{arniqa} & NR & $\phantom{-}0.533$ & $\phantom{-}0.489$ & $\phantom{-}0.103$ & $\phantom{-}0.190$ \\
PI~\citep{pi} & NR & $\phantom{-}0.419$ & $\phantom{-}0.374$ & $-0.101$ & $-0.039$ \\
NRQM~\citep{nrqm} & NR & $\phantom{-}0.403$ & $\phantom{-}0.355$ & $-0.074$ & $-0.004$ \\
PaQ-2-PiQ~\citep{paq2piq} & NR & $\phantom{-}0.401$ & $\phantom{-}0.283$ & $\phantom{-}0.148$ & $\phantom{-}0.228$ \\
ILNIQE~\citep{ilniqe} & NR & $\phantom{-}0.396$ & $\phantom{-}0.305$ & $-0.229$ & $-0.112$ \\
PIQE~\citep{piqe} & NR & $\phantom{-}0.380$ & $\phantom{-}0.363$ & $-0.061$ & $-0.046$ \\
BRISQUE~\citep{brisque} & NR & $\phantom{-}0.370$ & $\phantom{-}0.354$ & $\phantom{-}0.015$ & $\phantom{-}0.009$ \\
NIQE~\citep{niqe} & NR & $\phantom{-}0.360$ & $\phantom{-}0.327$ & $-0.077$ & $-0.044$ \\
NeuralSBS~\citep{neuralsbs} & NR & $-0.184$ & $-0.087$ & $\phantom{-}0.238$ & $\phantom{-}0.258$ \\
\bottomrule
\end{tabular}
\endgroup
\vspace{-15px}
\end{table*}

%% file: tab_metrics_on_content_types.tex
\begin{table*}[t]
\centering
\caption{Within-subset Spearman correlation (SRCC) with MOS across SR generators, input degradations, and upscaling factors. PSNR and TOPIQ-FR use a reference image (FR); Q-Align and distilled Q-ReAlign-mini do not (NR). Each correlation is computed separately on its indicated subset. In the \emph{SR method} rows, bold and underlined values mark the highest and lowest SRCC within each SR family and metric, respectively; they do not indicate statistical significance.}
\label{tab:metrics_on_content_types}
\begingroup
\footnotesize
\setlength{\tabcolsep}{5pt}
\renewcommand{\arraystretch}{0.92}
\begin{tabular}{@{}llcccc@{}}
\toprule
& & \multicolumn{4}{c}{SRCC with MOS} \\
\cmidrule(lr){3-6}
Factor & Subset & PSNR (FR) & TOPIQ-FR & Q-Align (NR) & \shortstack{Q-ReAlign-mini\\(distilled, NR)} \\
\midrule
\multicolumn{6}{@{}l}{\textit{Diffusion SR}} \\
Overall & All outputs & $0.257$ & $0.397$ & $0.415$ & $0.496$ \\
\addlinespace[2pt]
SR method & ResShift & $\mathbf{0.412}$ & $\mathbf{0.587}$ & $0.571$ & $0.613$ \\
 & SinSR & $0.342$ & $0.506$ & $0.428$ & $0.515$ \\
 & SeeSR & $0.301$ & $0.367$ & $0.301$ & $0.444$ \\
 & PASD & $0.239$ & $0.232$ & $0.223$ & $0.326$ \\
 & PASD-SDXL & $0.255$ & $0.393$ & $0.447$ & $0.468$ \\
 & SUPIR & $0.248$ & $0.407$ & $0.249$ & $0.356$ \\
 & DiT4SR & $\underline{0.183}$ & $\underline{0.152}$ & $\underline{0.178}$ & $\underline{0.313}$ \\
 & InvSR & $0.332$ & $0.405$ & $0.402$ & $0.527$ \\
 & CODSR & $0.196$ & $0.247$ & $0.287$ & $0.421$ \\
 & PassionSR & $0.317$ & $0.571$ & $\mathbf{0.714}$ & $\mathbf{0.666}$ \\
\addlinespace[2pt]
LR degradation & Blur & $0.251$ & $0.448$ & $0.473$ & $0.542$ \\
 & Noise & $0.268$ & $0.396$ & $0.403$ & $0.491$ \\
 & JPEG & $0.234$ & $0.363$ & $0.414$ & $0.495$ \\
 & Resized & $0.274$ & $0.393$ & $0.382$ & $0.466$ \\
\addlinespace[2pt]
Upscaling & $\times 2$ & $0.204$ & $0.245$ & $0.314$ & $0.442$ \\
 & $\times 4$ & $0.147$ & $0.399$ & $0.439$ & $0.505$ \\
\midrule
\multicolumn{6}{@{}l}{\textit{Non-diffusion SR}} \\
Overall & All outputs & $0.630$ & $0.811$ & $0.813$ & $0.729$ \\
\addlinespace[2pt]
SR method & Real-ESRGAN & $\underline{0.500}$ & $\underline{0.665}$ & $\underline{0.633}$ & $\mathbf{0.621}$ \\
 & RealSR & $0.606$ & $0.711$ & $\mathbf{0.769}$ & $\underline{0.306}$ \\
 & SwinIR & $0.504$ & $0.701$ & $0.633$ & $0.589$ \\
 & HAT & $\mathbf{0.631}$ & $\mathbf{0.715}$ & $0.738$ & $0.485$ \\
\addlinespace[2pt]
LR degradation & Blur & $0.403$ & $0.768$ & $0.737$ & $0.711$ \\
 & Noise & $0.711$ & $0.864$ & $0.842$ & $0.802$ \\
 & JPEG & $0.603$ & $0.831$ & $0.851$ & $0.757$ \\
 & Resized & $0.699$ & $0.858$ & $0.863$ & $0.804$ \\
\addlinespace[2pt]
Upscaling & $\times 2$ & $0.467$ & $0.689$ & $0.710$ & $0.724$ \\
 & $\times 4$ & $0.467$ & $0.669$ & $0.719$ & $0.578$ \\
\bottomrule
\end{tabular}
\endgroup
\end{table*}

%% file: tab_finetuned_metrics.tex
\begin{table*}[t]
\centering
\caption{Overall signed correlations with MOS for 10 fine-tuned baselines and one distilled student on non-diffusion and diffusion SR. Rows are ordered by non-diffusion SRCC within each type. Best and second-best results within each type and column are shown in bold and underlined, respectively. Values are rounded to three decimal places.}
\label{tab:metric_leaderboard_finetuned}
\begingroup
\footnotesize
\setlength{\tabcolsep}{4pt}
\renewcommand{\arraystretch}{0.88}
\begin{tabular}{@{}lccccc@{}}
\toprule
& & \multicolumn{2}{c}{Non-diffusion SR} & \multicolumn{2}{c}{Diffusion SR} \\
\cmidrule(lr){3-4}\cmidrule(lr){5-6}
Metric & Type & SRCC & PLCC & SRCC & PLCC \\
\midrule
LPIPS-VGG(FT)~\citep{lpips} & FR & $\phantom{-}\mathbf{0.778}$ & $\phantom{-}\mathbf{0.727}$ & $\phantom{-}0.267$ & $\phantom{-}0.306$ \\
LPIPS-AlexNet(FT)~\citep{lpips} & FR & $\phantom{-}\underline{0.752}$ & $\phantom{-}\underline{0.691}$ & $\phantom{-}0.294$ & $\phantom{-}0.331$ \\
AHIQ(FT)~\citep{ahiq} & FR & $\phantom{-}0.720$ & $\phantom{-}0.687$ & $\phantom{-}\underline{0.334}$ & $\phantom{-}\underline{0.388}$ \\
TOPIQ(FR)(FT)~\citep{topiq} & FR & $\phantom{-}0.694$ & $\phantom{-}0.680$ & $\phantom{-}\mathbf{0.436}$ & $\phantom{-}\mathbf{0.474}$ \\
PieAPP(FT)~\citep{pieapp} & FR & $\phantom{-}0.539$ & $\phantom{-}0.443$ & $\phantom{-}0.306$ & $\phantom{-}0.315$ \\
\midrule
Q-ReAlign-mini(distilled) & NR & $\phantom{-}\mathbf{0.729}$ & $\phantom{-}\mathbf{0.747}$ & $\phantom{-}\mathbf{0.496}$ & $\phantom{-}\mathbf{0.542}$ \\
Q-ReAlign-mini(FT) & NR & $\phantom{-}\underline{0.727}$ & $\phantom{-}\underline{0.674}$ & $\phantom{-}\underline{0.450}$ & $\phantom{-}\underline{0.470}$ \\
TOPIQ(NR)(FT)~\citep{topiq} & NR & $\phantom{-}0.637$ & $\phantom{-}0.594$ & $\phantom{-}0.369$ & $\phantom{-}0.412$ \\
MUSIQ(FT)~\citep{musiq} & NR & $\phantom{-}0.497$ & $\phantom{-}0.487$ & $\phantom{-}0.308$ & $\phantom{-}0.348$ \\
ARNIQA(FT)~\citep{arniqa} & NR & $\phantom{-}0.379$ & $\phantom{-}0.314$ & $-0.164$ & $-0.155$ \\
LIQE(FT)~\citep{liqe} & NR & $\phantom{-}0.309$ & $\phantom{-}0.144$ & $-0.028$ & $-0.016$ \\
\bottomrule
\end{tabular}
\endgroup
\end{table*}

%% file: tab_efficient_metric.tex
\begin{table}[t]
\centering
\caption{Accuracy and efficiency of the proposed student and its vision--language baselines. Accuracy is reported as SRCC/PLCC. SRIQA-Bench results use the 110-image validation split. Throughput and peak VRAM are measured on $512\times512$ inputs using an NVIDIA RTX A6000 in BF16.}
\label{tab:efficient_metric}
\begingroup
\footnotesize
\setlength{\tabcolsep}{4pt}
\renewcommand{\arraystretch}{0.95}
\begin{tabular}{@{}lcccccc@{}}
\toprule
Model & Params & VRAM & FPS & SRIQA-Bench & Diffusion & Non-diff. \\
\midrule
Q-ReAlign-mini & 0.8B & 2.3 GB & 3.76 & 0.479/0.504 & 0.341/0.393 & 0.683/0.597 \\
Q-ReAlign-lite & 4.0B & 10.7 GB & 2.69 & 0.475/0.519 & 0.405/0.458 & 0.772/0.719 \\
Q-ReAlign-pro & 9.0B & 19.2 GB & 2.65 & 0.428/0.488 & 0.431/0.469 & 0.760/0.716 \\
Q-Align & 8.2B & 16.0 GB & 5.39 & 0.624/0.677 & 0.415/0.487 & \textbf{0.813/0.758} \\
Ours & \textbf{0.34B} & \textbf{1.3 GB} & \textbf{6.37} & \textbf{0.755/0.794} & \textbf{0.496/0.542} & 0.729/0.747 \\
\bottomrule
\end{tabular}
\endgroup
\end{table}